\documentclass{article}
\usepackage{amsmath,graphicx,mlspconf}
\usepackage[utf8]{inputenc} 
\usepackage[english]{babel}
\usepackage{csquotes}
\usepackage[style = numeric-comp,
 giveninits=true , 
 natbib = true,
 doi=false, isbn=false,url=false,backend=biber]{biblatex}
\usepackage[nolist]{acronym}
\usepackage{amsmath}
\usepackage{amsfonts}
\usepackage{subcaption}
\usepackage{float}
\usepackage{color}

\usepackage{mdwlist}

\copyrightnotice{U.S.\ Government work not protected by U.S.\ copyright}

\copyrightnotice{978-1-5386-5477-4/18/\$31.00 {\copyright}2018 Crown}

\copyrightnotice{978-1-5386-5477-4/18/\$31.00 {\copyright}2018 European Union}

\copyrightnotice{978-1-5386-5477-4/18/\$31.00 {\copyright}2018 IEEE}

\toappear{2018 IEEE International Workshop on Machine Learning for Signal Processing, Sept.\ 17--20, 2018, Aalborg, Denmark}

\renewcommand{\vec}{\boldsymbol}

\DeclareMathOperator{\sgn}{sgn}

\title{A characterization of the Edge of Criticality \\ in Binary Echo State Networks}
\name{Pietro Verzelli$^1$, Lorenzo Livi$^2$, Cesare Alippi$^{1,3}$
}
\address{\small
$^1$Università della Svizzera Italiana, Lugano,Switzerland.$^2$ University of Exeter, Exeter,United Kingdom.$^3$ Politecnico di Milano, Milano,Italy.  }
\begin{document}
%\ninept
%

\maketitle

\begin{abstract}
Echo State Newtworks (ESNs) are simplified recurrent neural network models composed of a reservoir and a linear, trainable readout layer. The reservoir is tunable by some hyper-parameters that control the network behaviour. ESNs are known to be effective in solving tasks when configured on a region in (hyper-)parameter space called \emph{Edge of Criticality} (EoC), where the system is maximally sensitive to perturbations hence affecting its behaviour. In this paper, we propose  binary ESNs, which are architecturally equivalent to standard ESNs but consider binary activation functions and binary recurrent weights.
For these networks, we derive a closed-form expression for the EoC in the autonomous case and perform simulations in order to assess their behavior in the case of noisy neurons and in the presence of a signal. We propose a theoretical explanation for the fact that the variance of the input plays a major role in characterizing the EoC.
\end{abstract}
\section{Introduction}
\label{sec:intro}

\begin{acronym}
    \acro{bESN}{binary ESN}
	\acro{EoC}{Edge of Chaos}
	\acro{ESN}{Echo State Network}
	\acro{FIM}{Fisher Information Matrix}
	\acro{LSM}{Liquid State Machine}
	\acro{MFT}{Mean Field Theory}
	\acro{ML}{Machine Learning}
	\acro{PDF}{Probability Density Function}
	\acro{RBN}{Random Boolean Network}
	\acro{RNN}{Recurrent Neural Network}
	\acro{RP}{Recurrency Plot}
\end{acronym}

% Select one of the four copyright notices below and put in the text field below the first column (only required for the camera paper submission). \textbf{Please pay attention to that most authors should use Copyright notice 4}.

% \noindent{\bf Copyright notice 1:}\\
% For papers in which all authors are employed by the US government, the copyright notice is:\\
% {U.S.\ Government work not protected by U.S.\ copyright}
% \\[2ex]
% \noindent{\bf Copyright notice 2:}\\
% For papers in which all authors are employed by a Crown government (UK, Canada, and Australia), the copyright notice is:\\
% {978-1-5386-5477-4/18/\$31.00 {\copyright}2018 Crown}
% \\[2ex]
% \noindent{\bf Copyright notice 3:}\\
% For papers in which all authors are employed by the European Union, the copyright notice is:\\
% {978-1-5386-5477-4/18/\$31.00 {\copyright}2018 European Union}
% \\[2ex]
% \noindent{\bf Copyright notice 4:}\\
% For all other papers the copyright notice is:\\
% {978-1-5386-5477-4/18/\$31.00 {\copyright}2018 IEEE}

\acp{ESN} \cite{jaeger2001echo} maximize predictive performance on the Edge of Criticality or \ac{EoC}, which is a region in parameter space where the system is maximally sensitive to perturbations \cite{legenstein2007edge}.
To date, a complete theoretical understanding for the \ac{EoC} is missing for input-driven \acp{ESN} without mean-field assumptions.
This performance maximization at the \ac{EoC} is known to be common to many dynamical systems, even really simple ones \cite{langton1990computation}. 
This property has lead many researcher to study these models in order to explore the main features associated with the transition to chaos.
In this direction, \acp{RBN} \cite{kauffman1969metabolic} are well studied networks. Their chaotic behavior is well understood (
\cite{derrida1986random, wang2011fisher, farkhooi2017complete}), though their applicability is scarce due to the that they only input binary signals.

In this paper, we propose \acp{bESN}. The architecture is equivalent to standard \acp{ESN} but simplified as they consider binary activation functions and binary weights for the recurrent connections. To the best of out knowledge, this architecture has never been investigated before.
\acp{bESN} share some similarities also with a particular form of \acp{RBN}, called Random Threshold Networks \cite{rohlf2002criticality} (that are based on the same idea of summing the input of neurons).
We derive a closed-form expression to determine the \ac{EoC} in autonomous \acp{bESN} (i.e., the network is not driven by signals) and perform simulations to assess the behavior of \acp{bESN} both in the autonomous and non-autonomous case. We experimentally assess the quality of our theoretical prediction regarding the onset of chaos in \acp{bESN} in the autonomous case. Results show perfect agreement with the theory. Then, in order to asses the network stability, we analyze the impact of noise on the neuron outputs on the onset of chaos. Our findings suggests that the chaotic region expands linearly with the noise intensity when the mean degree (i.e., the average number of links a neuron has) is high enough.
We also study the \ac{EoC} for \acp{bESN} driven by (continuous) signals, discussing how our findings could be generalized considering the signal gain as an hyperparameter.
This work sets also in the context of reducing model complexity, in which binarization plays an important role both for theoretical aspects \cite{baldassi2016learning} and applied perspectives \cite{cheng2018model}, since it would considerably reduce required hardware resources and speed-up training algorithms.

\section{Background material} \label{sec:BG}
    \subsection{Echo State Networks} \label{subsec:ESN}
    An \ac{ESN} is a discrete-time non-linear system with feedback, whose model reads:
    \begin{align}
    \vec{x}[n+1] =& f\left(
    \vec{W} \vec{x}[n] + 
    \vec{W}_i^r \vec{u}[n+1] + 
    \vec{W}_o^r  \vec{y}[n]
    \right) \label{eqn:ESN_x}
    \\
    \vec{y}[n+1] = & g\left(
    \vec{W}_i^o \vec{u}[n+1] + 
    \vec{W}_r^o \vec{x}[n+1]
    \right) \label{eqn:ESN_y}
    \end{align}
    
    An \ac{ESN} consists of a \emph{reservoir} of $N$ neurons characterized by a non-linear transfer function $f(\cdot)$. At time $n$ the network is driven by the input $\vec{u}[n] \in \mathbb{R}^{N_i}$ and produces an output $\vec{y}[n] \in \mathbb{R}^{N_o}$, $N_i$ and $N_o$ being the dimensions of the input and output vectors, respectively.
    
    The weight matrices $\vec{W}\!\in\! \mathbb{R}^{N \times N}$ (reservoir internal connections), $\vec{W}_i^r \!\in\! \mathbb{R}^{N \times N_i}$ (input-to-reservoir connections) and $\vec{W}_o^r \!\in\! \mathbb{R}^{N \times N_o}$ (output-to-reservoir feedback connections) contain values in the $[-1,1]$ interval drawn from a uniform (or sometimes Gaussian) distribution.
    The output weight matrices  $\vec{W}_i^o \!\in\! \mathbb{R}^{N_o \times N_i}$ and $\vec{W}_r^o  \!\in\! \mathbb{R}^{N_o \times N_r}$, connecting reservoir and input to the output, represent the readout layer of the network. Activation functions $f(\cdot)$ and $g(\cdot)$ (applied component-wise) are typically implemented as a sigmoidal ($\tanh$) and identity function, respectively. Training requires solving a regularized least-square problem \cite{jaeger2001echo}.
    
    Various empirical results suggest that \acp{ESN} achieve the highest expressive power, i.e., the ability to provide optimal performances, exactly when configured on the edge of the transition between a ordered and chaotic regime (e.g., see 
    \cite{jaeger2001echo, yildiz2012re,livi2017determination, rivkind2017local, legenstein2007edge,rajan2010stimulus}). %, sussillo2009generating
    Once the network operates on the edge -or in proximity to - it achieves the highest memory capacity (storage of past information) and accuracy prediction, compatible with the network architecture.
    For determining the edge of chaos, one usually resorts to computing the \emph{maximum Lyapunov exponent} \cite{gallicchio2017echo} or identify parameter configuration maximizing the \emph{Fisher information} \cite{livi2017determination}.
\subsection{Random Boolean Networks}\label{subsec:RBNs}
    \acp{RBN} where first proposed in \cite{kauffman1969metabolic} as a model for the gene regulatory mechanism. The model consists of $N$ variables $\sigma_i \in \{0,1\}$ -- sometimes called \emph{spins} -- whose time-evolution is given by $\sigma_i[n+1] = f_i(\sigma_{i_1}[n], \sigma_{i_2}[n], \dots , \sigma_{i_K}[n])$, where each $f_i \in \{ 0,1 \}$ is a Boolean function of $K$ variables, representing the $K$ incoming links to the $i$-th element of the net. There exists $2^{2^K}$ possible Boolean functions of $K$ variables. The output of $f_i$ is randomly chosen to be $1$ with probability $r$ and $0$ with probability $1-r$, so that usually one refers to $r$ as the \emph{bias} of the network. 
    
    \acp{RBN} show two distinct regimes, depending on both the value of $K$ and $r$: a phase in which the network assumes a \emph{chaotic} behaviour and a phase (sometimes called \emph{frozen}) in which the network rapidly collapses to a stable state. In \cite{derrida1986random}, the authors justify this behavior by studying the evolution of the (Hamming) distance of two (initially different) configurations. There, they derive the following formula for the onset of the chaos: $K > K_c = \frac{1}{2r(1-r)}$.
\section{Binary echo state network} \label{sec:bESN}
    In this section, we introduce \acp{bESN} and study the dynamics of a reservoir similar to \eqref{eqn:ESN_x} constituted of binary neurons $x_i\! \in\! \{ -1, +1 \}$ for $ i \!=\! 1,2,...,N $ and binary weights $W_{ij} \in \{ -1, 0, +1 \}$  for $ i,j \!=\! 1,2,...,N$ (the zero value accounts for the fact that two neurons may not be linked). For simplicity, we will not consider feedback connection (i.e., $\vec{W}^r_o =0$) . The \acp{bESN} system model simplifies as:
    \begin{align}
        &x_i[n+1] = \sgn( S_i[n] ), \label{eqn:bESN_x}
        \\
        &S_i[n] := \sum_{j=1}^{N}W_{ij}x_j[n] + u[t]. \label{eqn:bESN_S}
    \end{align}
    $u[t]$ is the input signal, which we consider to be unfiltered ($(W_i^r)=\mathbf{1}$, i.e., the all-ones vector), for simplicity.
    When $u[n] = 0$ for every $n$ (i.e., there is no input), we say that the system is \emph{autonomous}. The study of the autonomous system plays an important role, since it allows us to investigate analytically the network dynamics and its properties.
    %We can think of a reservoir as a \emph{directed weighted graph}, completely defined by its weights matrix $\vec{W}\!=(\!W_{ij})$, which acts as a weighted adjacency matrix. 
    Reservoir connections $\vec{W}\!=(\!W_{ij})$ are instantiated according to the Erdős–Rényi model where each link $W_{ij}$ is created with probability $\alpha$. If the link is generated, the weight value is set to $1$ with probability $p$ or $-1$ with probability $1 -p$.
    
    The proposed \ac{bESN} model is controlled by three hyperparameters: (1) $N$, the \emph{number of neurons} in the network; (2) $\langle k \rangle := \alpha N$, the \emph{mean degree} of the network; (3) $d := p - \frac{1}{2}$, the \emph{asymmetry} in the weights values.
    These hyperparameters are related to $\alpha$ and $p$, although they are easier to understand: in fact, $\langle k \rangle $ has a natural interpretation in terms of mean neuron degree that does not depend on the network size $N$. The choice of using $d$ is due to the symmetry of the model around the zero value and to the fact that, with this choice, a positive (negative) value of the hyperparameter accounts for majority of positive (negative) weights. Note that $\langle k \rangle$ can vary \emph{continuously} from $0$ to $N$ and  $d\in(-\frac{1}{2},\frac{1}{2})$. 
    %although it cannot be considered as a probability. 
    A similar model was proposed in \cite{rohlf2002criticality}, but in their work the weights assume a positive or negative value with equal probability, i.e., their model corresponds to ours in the $p = 0$ case.

    \subsection{Edge of chaos in binary ESN}\label{subsec:EoC_theory}
    % where $\sigma^2 = K(1-4d^2)$ is the variance of the activation.
    Here, we study two networks with the same weight matrix $\vec{W}$, that are in the states $\vec{x}[n]$ and $\vec{x}'[n]$, the latter refers to the perturbed network and differs only in one neuron whose state is flipped.
    The goal is here to understand under which conditions the time evolution of the perturbed network differs from the original one, i.e., whether the perturbation will spread and significantly impact the network behavior or not.
    
    For $N \!\to \! \infty $, the fraction of positive-valued neurons is equal to the probability for a neuron of being positive, namely $P_+\!=\!p$, while the fraction of negative neurons is $P_- \!=\! q \!=\! 1-p$. By comparing the original network with the perturbed one, the probability that the flipped neuron will have an influence on a neuron connected to it will be given by two terms: the probability that neuron state is positive $P_+$ multiplied by the probability of switching from positive to negative $\pi_{+-}$, plus an analogous terms accounting for the negative part ($P_-$ and $\pi_{-+}$ respectively. Assuming that $\pi_{+-}\!=\!P_-\!=\!q$ (i.e., that the probability of turning negative from positive is equal to the probability of being negative) and, analogously, that $\pi_{-+}\!=\!P_+\!=\!p$ (that can be seen as a formulation of the \emph{annealed approximation} introduced in \cite{derrida1986random}), one obtains:
    \begin{equation}\label{eqn:proba_bESN}
    \begin{aligned}
        P_+ \cdot \pi_{+-} +  P_- \cdot \pi_{-+} &= pq+qp = 2p(1-p)
    \end{aligned}
    \end{equation}
    We now define $k_{O}$ and $k_{I}$ as the mean \emph{out-degree} and \emph{in-degree} of a neuron, respectively. Since a single neuron has influence on $k_{O}$ neurons, the expected number of changes is given by $2p(1-p)k_O$, to which one has to add the fact that at least one neuron has changed due to the flip. Therefore, if this number is bigger than half of the mean incoming links of a neuron, i.e., $k_I/2$, then the perturbation will dominate the network dynamics and will propagate.
    Since in an Erd\"os–R\'enyi graph $k_I \!= \!k_O \!=\! \langle k \rangle$, we obtain the following condition for the onset of chaos:
        \begin{equation} \label{eqn:condition}
            1+ 2p(1-p)\langle k \rangle > \frac{\langle k \rangle}{2}
        \end{equation}
    which using $d := p - 1/2$ can be rewritten as:
        \begin{equation}\label{eqn:bESN_EoC}
            \langle k \rangle < k_c := \frac{1}{2d^2}
        \end{equation}
    Note that the mean degree $\langle k \rangle$ in \eqref{eqn:bESN_EoC} plays a ``stabilizing'' role (i.e., the higher the degree, the larger the magnitude of $d$ required for chaos), as opposed to \acp{RBN}, where increasing the mean degree leads towards a chaotic region.

\section{Experiments}

%In Sec.~\ref{subsec:EoC}, we assess the agreement of our theoretical expression for the EoC with simulations. Initially, we consider an autonomous bESN.
%In Sec.~\ref{subsec:perturbation} we analyze the effects of perturbations on network trajectories.
%In Sec.~\ref{subsec:exp_noise} we take into account the effect of white Gaussian noise on the theoretical predictions of the EoC; in Sec.~\ref{subsec:exp_signal} we consider periodic inputs and signals generated from a random walk (Brownian motion) with increasing degree of autocorrelation for the increments.

    \subsection{Edge of chaos}
    \label{subsec:EoC}
    In order to assess the agreement between the prediction given by Eq. \ref{eqn:bESN_EoC} and experimental results, we conducted an exploration of the parameter space. Here, we exploit the fact that our neurons assume binary states only and consider their \emph{Shannon entropy} $H$ as an indicator for the transition to chaos. The entropy $H$ was computed considering a time average $\overline{H}$,
    \begin{equation}\label{eqn:entropy_mean}
        \overline{H} := 
        \frac{1}{T-t_0} \sum^{T}_{n = t_0} H(\vec{x}[n])
    \end{equation}
    where the entropy of a configuration $\vec{x}$ is estimated as $H(\vec{x}) := -\rho(\vec{x}) \log (\rho(\vec{x})) - (1-\rho(\vec{x})) \log (1 -\rho(\vec{x}))$, in which $\rho(\vec{x})$ is the number of neurons whose state is $+1$ and $(1 -\rho(\vec{x})$ the number of neurons with state $-1$.
    We expect Eq.\eqref{eqn:entropy_mean} to be be almost zero in the frozen regime and almost one in the chaotic one, with a sharp region of intermediate values that we consider to be the edge of chaos.
    
\begin{figure}[]
        \centering
        \includegraphics[width=.4\textwidth]{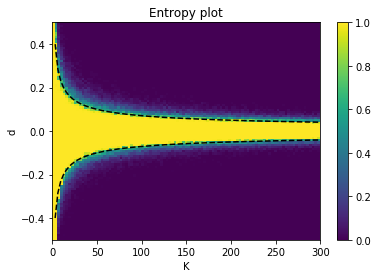}
        \caption{Values of $\overline{H}$ for different configurations of the asymmetry and the mean degree. The experiment shows a good agreement with the predicted \ac{EoC} region (dashed line), where we observe an abrupt change of the entropy from $0$ to $1$.}
        \label{fig:entropy}
    \end{figure}

    In order to explore the parameter space, we run a series of simulations using a network of $N\!=\!1000$ neurons with different random initial conditions and connection matrices $\mathbf{W}$, generated using specific values of $\langle k \rangle$ and $d$.
    In Eq. \ref{eqn:entropy_mean}, we used $T=300$ time-steps and $t_0=100$ accounting for an initial transient from the initial state to a stationary condition.
    Results are showed in Fig.~\ref{fig:entropy} and demonstrate almost perfect agreement with the theoretical result \eqref{eqn:bESN_EoC}.

    \subsection{Effects of perturbations on state evolution}
    \label{subsec:perturbation}
    
    In order to assess the effect of chaos on the network behavior, we compare the evolution of a \ac{bESN} instantiated with weight matrix $\vec{W}$ but different initial conditions. Starting from a random initial condition, we generated $50$ additional initial conditions by flipping the state of a single neuron (as in Sec.~\ref{subsec:EoC_theory}). Here, an initial condition $\vec{x}_0$ is randomly generated with a biased probability $c=0.6$ for a neuron to assume a positive value.
    %This becomes necessary as the network in the frozen phase could reach two different stable states -- one with almost all positive neurons and another one with almost all negative neurons -- making the following analysis impossible. Finally, we note that the value of $c$ has no impact on the \ac{EoC}.
    We let the original and perturbed networks evolve, and take into account the (normalized) Hamming distance, $D_H$, between trajectories.
    
\begin{figure*} 
    \centering
    \begin{subfigure}[b]{0.65\textwidth}
        \includegraphics[width=\textwidth]{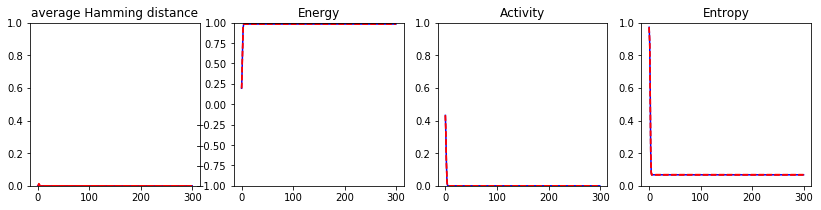}
        \caption{$d=0.25$}
        \label{fig:025}
    \end{subfigure}
    \begin{subfigure}[b]{0.65\textwidth}
        \includegraphics[width=\textwidth]{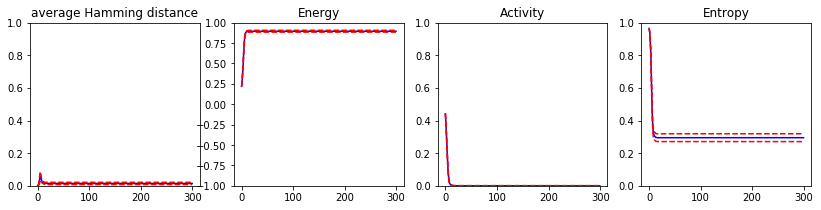}
        \caption{$d=0.184$}
        \label{fig:0184}
    \end{subfigure}
    
    \begin{subfigure}[b]{0.65\textwidth}
        \includegraphics[width=\textwidth]{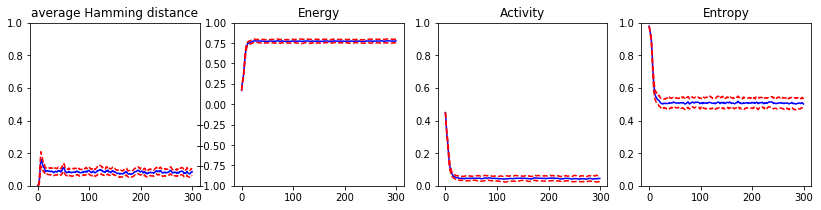}
        \caption{$d=0.157$}
        \label{fig:0157}
    \end{subfigure}
    
    \begin{subfigure}[b]{0.65\textwidth}
        \includegraphics[width=\textwidth]{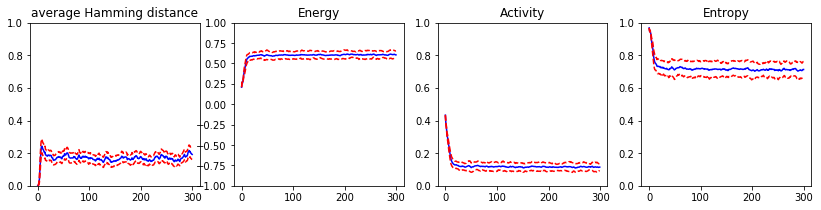}
        \caption{$d=0.144$}
        \label{fig:0144}
    \end{subfigure}

    \begin{subfigure}[b]{0.65\textwidth}
        \includegraphics[width=\textwidth]{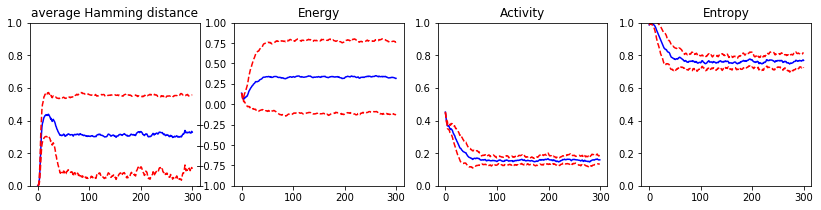}
        \caption{$d=0.131$}
        \label{fig:0131}
    \end{subfigure}
    
    \begin{subfigure}[b]{0.65\textwidth}
        \includegraphics[width=\textwidth]{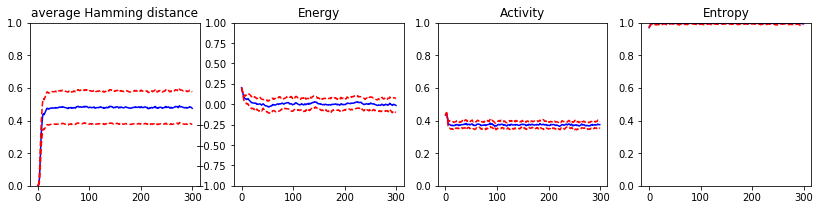}
        \caption{$d=0.105$}
        \label{fig:0105}
    \end{subfigure}
    
    \caption{Mean values of the Hamming distance, Energy, Activity and Entropy of the $50$ perturbed networks, with $N=1000$ and $K=22$ for selected values of $d$ (see \ref{subsec:perturbation}). The $x$-axis represents time. The values of the quantities are plotted in blue, while the dashed red lines show the variance. The predicted system should turn chaotic for $d < 1/\sqrt{2 \langle k \rangle} \approx 0.15$ , according to the theoretical formula.
    }
    \label{fig:perturbations}
\end{figure*}

    Results are summarized in Fig. \ref{fig:perturbations}.
    We observe that, in the ordered phase, perturbations on the initial state have no effect on the network evolution and the Hamming distance of the perturbed trajectory from the original one is zero. As $d$ decreases (i.e., the networks approach the chaotic regime), we observe how the Hamming distance significantly increases, leading to chaos. Note that the maximum value achievable by the (normalized) Hamming distance is $0.5$, corresponding to the distance of two random binary vectors (a larger distance would imply a negative correlation).
    In the same set of figures, we show three additional indicators, called \emph{Energy}, \emph{Activity}, and \emph{Entropy}.
    The mean \emph{Energy}, defined as $E(\vec{x}[n]) := \frac{1}{N}\sum_{i=1}^N x_i[n]$, quantifies the average number of positive and negative neurons.
    In the frozen phase, the network almost instantly evolves towards values close to $1$ (cfr. the role of $c$, discussed above), and then rapidly decreases to $0$, which is the expected value in the chaotic phase. The mean \emph{Activity} of network at time-step $n$ is defined as the (normalized) Hamming distance of the current state w.r.t the previous one, $A(\vec{x}[n]) = D_H(\vec{x}[n], \vec{x}[n-1])$, i.e., the number of neurons that changed their states in one step. As expected, networks operating in chaotic regimes are characterized by an elevated activity. Lastly, we plot the evolution of the \emph{Entropy} \eqref{eqn:entropy_mean} over time. As expected from the theory, transitioning to a chaotic regime is signaled by a sharp increase of entropy.
    \subsection{Impact of noise in bESN edge of chaos}
    \label{subsec:exp_noise}
    Here, we study how the \ac{EoC} is influenced when considering an independent noise term for each neuron, $x_i[n+1] = \sgn \! \left( S_i[n] + \nu \!\cdot\! \langle k \rangle\! \cdot\! \xi_i[n] \right)$, where $\nu$ is the \emph{noise gain}, $\xi_i\sim\mathcal{N}(0, 1)$, and $S_i$ is the same as in Eq.\eqref{eqn:bESN_S}.
    The choice of scaling the noise with $\langle k \rangle$ was made to account for the fact that the network stability increases with it, as we discuss below.
    
\begin{figure}[ht!]
        \centering
        \includegraphics[width=.4\textwidth]{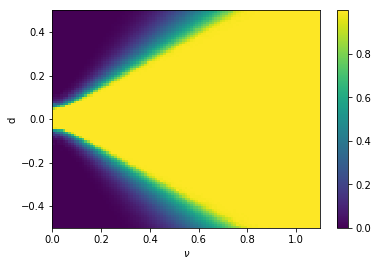}
        \caption{Values of $\overline{H}$ for different configurations of $d$ and the $\nu$. Note that $\nu$ is multiplied by the mean degree, which is here fixed to $\langle k \rangle=200$.}
        \label{fig:noiseVSd_k200}
    \end{figure}

\begin{figure}[ht!]
        \centering
        \includegraphics[width=.4\textwidth]{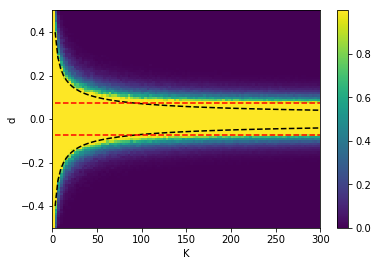}
        \caption{The same experiment of Fig.\ref{fig:entropy}, but with the presence of a noise term with $\nu = 0.1$ (multiplied by the mean degree, so that it increases along the $x$-axis). Note how for higher degree the chaos region is constant (the predicted value is the red dashed line), deviating from the autonomous-case prediction (red dashed line). 
        }
        \label{fig:entropy_noise}
    \end{figure}

    To explore the dependency from $\nu$, we ran an experiment where we fixed $\langle k \rangle$ and plotted $\nu$ versus $d$.
    Results are shown in Fig.~\ref{fig:noiseVSd_k200}. We can recognize three regimes: (1) for low noise values, the chaotic region remains almost constant; (2) for intermediate values, the chaotic region linearly expands with the noise intensity up until (3) there is only chaos. We repeated the experiments with different values of $\langle k \rangle$ (not shown) and they all confirm the same linear expansion of the chaotic region (in units of $\langle k \rangle$).
    To verify this fact for a wider range of $\langle k \rangle$, we repeated the experiment in Fig.~\ref{fig:entropy} with noise intensity $\nu = 0.1$. It is possible to observe in Fig.~\ref{fig:entropy_noise} how the \ac{EoC} maintains its shape for lower values of $\langle k \rangle$, while for higher average degrees it deviates from the theoretical prediction and the chaotic region depends on $d$ only.
        \begin{figure}[ht!]
        \centering
        \includegraphics[width=.4\textwidth]{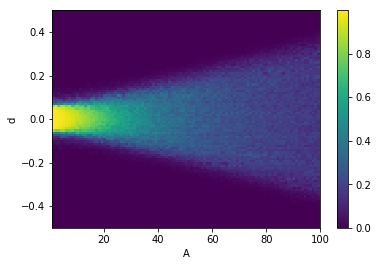}
        \caption{Network driven by white noise. Values of $\overline{H}$ for different configurations of $A$ and $d$. The mean degree was fixed to $\langle k \rangle=150$.}
        \label{fig:Ampiezza_WN}
    \end{figure}
    \begin{figure}[ht!]
        \centering
        \includegraphics[width=.4\textwidth]{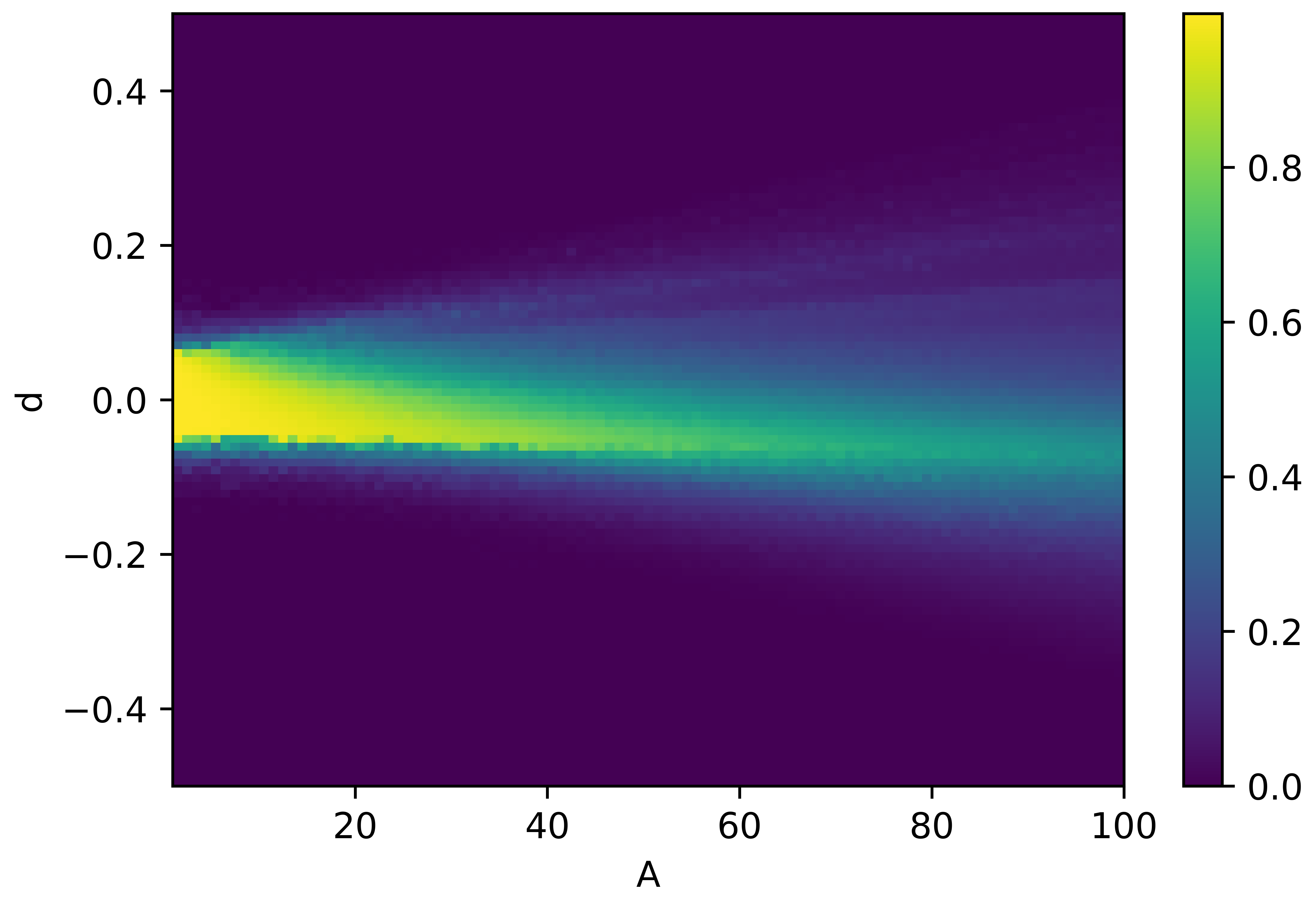}
        \caption{Network driven with the sum of three sinusoids. Values of $\overline{H}$ for different configurations of $A$ and $d$. The mean degree was fixed to $\langle k \rangle=150$.}
        \label{fig:Ampiezza_3seni.pdf}
    \end{figure}
    We explain this fact as follows. Neurons can only assume $1$ or $-1$ values. 
    %if we study the distribution of the positive neurons we will automatically know the total distribution. 
    The probability of a neuron having $j$ positive inputs is then
    $P_k(j) = {\binom{k}{j}} p^j q^{k-j}$ where $k$ is its in-degree. If we consider that $j \!= \!\frac{k+s}{2}$, where $s$ is the value of the \emph{sum of the positive and negative inputs} (whose sign determines the value of the neuron), then we obtain $s \!= \!2j-k$. The expectation of $j$ is $\langle j \rangle \! = \!pk$, so that the expectation of $s$ and its variance are:
    \begin{gather}
        \langle s \rangle = k(p-q) = 2kd \\
        \langle (s-\langle s\rangle)^2\rangle = 4kpq = k(1-4d^2)
    \end{gather}
    Note that these values are related to a single neuron. For a general understanding of the network behavior, one simply uses $\langle k \rangle$ instead of $k$ in the expressions above, so that it is possible to consider $\sigma^2 := \langle k \rangle(1-4d^2)$ as a \emph{mean-field variance} of the total inputs to neurons. The impact of the noise on the network can then be studied considering the ratio between $\nu$ and $\sigma^2$.
    As previously discussed, the noise expands the chaos region linearly with its $\nu$.
    
    The noise we are considering has a standard deviation $\theta_k = \nu \!\cdot\! \langle k \rangle$. This leads us to a formula for the chaotic region 
    which, for $\langle k \rangle \gg 1$, is $|d| < a\cdot \nu+b$. This relation, as shown in Fig.~\ref{fig:entropy_noise}, does not depend on $\langle k \rangle$. As such, having Gaussian noise with standard deviation $\theta$, the formula is $|d| < a\cdot\frac{\theta}{\langle k \rangle}$, or in terms of $\langle k \rangle$, we have $\langle k \rangle < k_c^{\textup{noise}}:=\frac{a\theta}{|d|}$.
    In our experiments, constant $a$ was determined as $a \approx 0.65$.
    
    \subsection{Impact of a signal}
    \label{subsec:exp_signal}

    As for the noise, the magnitude of the signal should have a major role in the \ac{EoC}, but this time the chaotic region should reduce instead of expanding, since the signal is known to suppress chaos in certain conditions \cite{rajan2010stimulus}. The signal introduces a correlation among neurons, which makes the annealed approximation ineffective.
    We drive the network with the signal as in \eqref{eqn:bESN_S}, but we scale $u[n]$ with a \emph{gain} factor $A$, since we are interested in its usage as an hyperparameter and not in relation with $\langle k \rangle $. 
    We initially feed the network with white noise (note that this is different from what we did in \ref{subsec:exp_noise}, since in this case the noise is the same for each neuron): from Fig.~\ref{fig:Ampiezza_WN} one can observe how the chaotic region rapidly shrinks as $A$ increases, but a region with an intermediate value of entropy expands (linearly). This is due to the fact the signal prevents the system from collapsing in a stable state, keeping the entropy above zero.
    
    In Fig.\ref{fig:Ampiezza_3seni.pdf} we show the results obtained for the normalized sum of three sines with  incommensurable frequencies (repeated also with different numbers of sinudoids, not shown). Again we note how the chaotic region shrinks as $A$ increases, with the appearance of the region characterized by intermediate entropy values which, instead, expands.

\section{Conclusions}
\label{sec:conclusion}

The binary \ac{ESN} model herein introduced is in principle similar to regular \acp{ESN}. However, its simplicity permits a theoretical analysis of some important aspects of the transition to chaos. The expression we derived here for the autonomous case perfectly matches the experimental results. Our analysis of the noise applied to neuron activations showed how the network stability increases linearly with the mean degree of recurrent connections.
The effects of input signals on the network dynamics are more complex to understand, since they introduce correlations among neurons.
Our analysis partially explained the role that the signal magnitude and the mean degree play in shaping the $\ac{EoC}$ of the non-autonomous case.

\printbibliography[heading=bibintoc]

\vfill
\pagebreak

\end{document}